# Hyperspectral images classification and Dimensionality Reduction using Homogeneity feature and mutual information

[1] Hasna Nhaila*, [2] Maria Merzouqi, [3] Elkebir Sarhrouni, [4] Ahmed Hammouch

Electrical Engineering Research Laboratory, ENSET

Mohammed V University

Rabat, Morocco

[1] hasnaa.nhaila@gmail.com, [2] merzouqimaria@gmail.com, [3] sarhrouni436@yahoo.fr, [4] hammouch_a@yahoo.com

*Abstract* - The Hyperspectral image (HSI) contains several hundred bands of the same region called the Ground Truth (GT). The bands are taken in juxtaposed frequencies, but some of them are noisily measured or contain no information. For the classification, the selection of bands, affects significantly the results of classification, in fact, using a subset of relevant bands, these results can be better than those obtained using all bands, from which the need to reduce the dimensionality of the HSI. In this paper, a categorization of dimensionality reduction methods, according to the generation process, is presented. Furthermore, we reproduce an algorithm based on mutual information (MI) to reduce dimensionality by features selection and we introduce an algorithm using mutual information and homogeneity. The two schemas are a filter strategy. Finally, to validate this, we consider the case study AVIRIS HSI 92AV3C.

## I. INTRODUCTION

In the area of HSI classification, an important question that often arises is the problem of having too many attributes. In other words, the measured attributes are not necessarily all needed for an accurate discrimination and the use of the entire set of these attributes can lead to a poor classification model. Indeed, hyperspectral data are expressed in high-dimensional spaces, and in directions containing various noises. This explains the curse of dimensionality [1]. This problem is compounded by the fact that many attributes can be either irrelevant or redundant because they don't add anything new to the result of prior classification.

In many applications, such as remote sensing with hyperspectral images, reduce the number of irrelevant or redundant attributes decreases significantly the execution time of a learning algorithm. So the problem is to find the right group of bands to reduce the dimensionality and classify the images.

## II. CATEGORIZATION OF DIMENSIONALITY REDUCTION METHODS ACCORDING TO ATTRIBUTES GENERATION PROCESS

According to generating attributes [2], the dimensionality reduction can be done either by:

- Attributes Extraction where we transform the vectors of data.
- Attributes Selection without transformation of the data vectors.
- Selection followed by extraction of attributes.

### A. Dimensionality Reduction by Attributes Selection

The idea of these methods is to find a subset of attributes having less wide than the initial one.

The selection of attributes (also known as subset selection) is also described as a process commonly used in the pretreatment before the classification step, in which a subset of variables (or attributes), from the available data, is selected for the application of a learning algorithm. The best subset contains the minimum number of dimensions that can lead to higher classification accuracy, we discard other irrelevant dimension. This is an important pre-processing step and it's one of two ways to avoid the curse of dimensionality (of course the other one is the attribute extraction) [2].

The algorithms for dimensionality reduction generally include four basic steps [3], see Figure 1:

1) *A procedure for generating the next candidate,*
2) *An evaluation fonction to evaluate the current subset,*
3) *A stopping criterion to decide when to stop the search,*
4) *A validation process to choose if we keep the subset or not.*

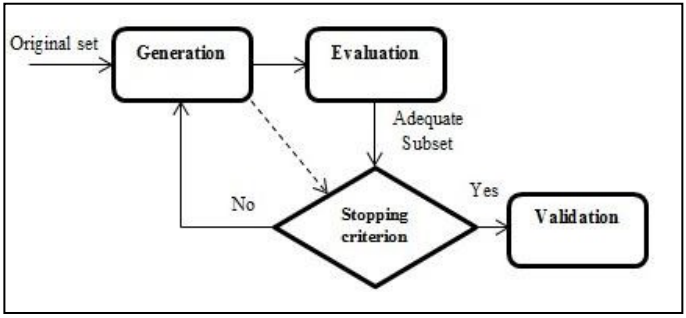

Fig.1. Attributes Selection Process with validation [3]

*B. Dimensionality Reduction by Attributes Extraction*

This approach consists in reducing the dimensionality of attributes by transformation of data, in fact, the original space is projected into a subspace of lower dimension which preserves most of the information.
So the idea is to transform the measurements, by linear or non-linear functions, in a preprocessing step, which generally entails a reduced set of derived variables which will used as the inputs of a classifier [4]. See figure 2.

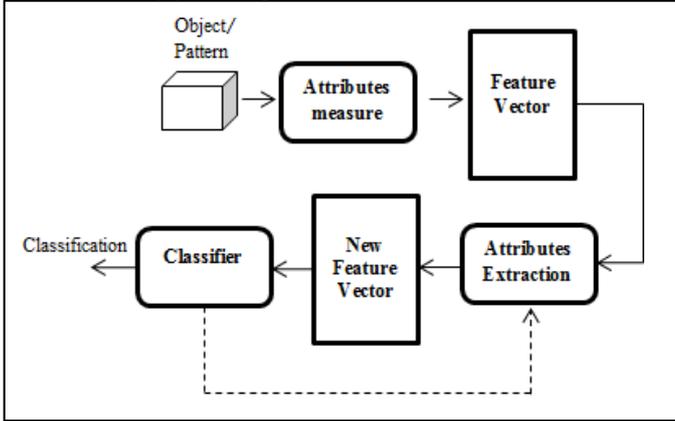

Fig. 2. Attributes Extraction Process, the Feedback corresponds to the Iterative research

In the next sections of this article, we will focus on the dimensionality reduction with attributes selection.

### III. DIMENTIONALITY REDUCTION BY FEATURES SELECTION USING MUTUAL INFORMATION

The aim of this section is to provide an application that illustrates the filter approach to reduce dimensionality of HSI. This is a schema that uses a measure of information, which is mutual information, and proceeds by sequential selection "forward" [5].
We start with a brief reminder of the mutual information, then we will focus on its application in HSI dimensionality reduction algorithm.

*A. Definition and measure of the mutual information*

It is a statistical measure of mutual information between the reference (ground truth map) that we note A, and each band noted B.

$$I(A,B) = \sum \log_2 p(A,B) \frac{p(A,B)}{p(A).p(B)} \quad (1)$$

We consider ground truth map and bands as random variables and we calculate their interdependence as illustrated in figure4 in the section of results and discussion.

*B. Filter approach using mutual information*

In this section we reproduce a "filter approach" based on mutual information.
The basic idea is: the band that has the largest value of mutual information with the ground truth, is a good approximation of it. Thus the subset of suitable bands is the one that generates the closest estimation to the ground truth GT.
We generate the current estimation by the average of the last estimation of the GT with the candidate band [5].

The first selecting processes "algorithm1" is as follows:
1) Order the bands according to decreasing value of their mutual information with the GT.
2) Initialize all the bands selected by the band that have the largest mutual information value with the GT.
3) Now, we build an approximation of the GT, denoted GT_est.
4) Calculate MI: MI (GT_est, GT). The last added band must increase the final value of IM (GT_est, GT), otherwise, it will be rejected from the choices.
5) Finally, we introduce a threshold to control the permitted redundancy.

### IV. DIMENTIONALITY REDUCTION BY FEATURES SELECTION USING HOMOGENEITÉ AND MUTUAL INFORMATION

In this method, we propose to combine the spectral information calculated by the IM, with the inter spatial information represented by the homogeneity that characterizes the texture bands.

*A. Definition and measurement of the homogeneity*

Texture analysis refers to the characterization of regions in an image by their texture content. Some of the most commonly used texture measures are derived from the Grey Level Co-occurrence Matrix (GLCM). In our case we will use the Homogeneity.
This statistic measures the closeness of the distribution of elements in the GLCM to the GLCM diagonal.

$$Homogeneity = \sum_i \sum_j \frac{1}{1+(i-j)^2} * \hat{P}(i,j) \quad (2)$$

The homogeneity value increases if the pixels values of the images are more similar. It has maximum value when all elements are same.

*B. Proposed Algorithm*

In this algorithm2, we will subjoin, to the mutual information used previously in algorithm1, one of the image texture characteristics which is the homogeneity extracted from the Grey Level Co-occurrence Matrix (GLCM).
So the selection process is :

1) Calculate the Grey Level Co-occurrence Matrix of the bandes and the GT.
2) Extract the texture feature of bands which is homogeinité. See figure 5 in the next section.
3) Order the bands according to decreasing value of their homogeneity values.
4) Initialize all the bands selected by the band that have the largest homogeneity value with the GT.
6) Construct an approximation of the GT, denoted GT_est.
7) Calculate MI: MI (GT_est, GT). The last added band must increase the final value of IM (GT_est, GT), otherwise, it will be rejected from the choices.
8) Finally, we introduce a threshold to control the permitted redundancy.

## V. RESULTS AND DISCUSSION

*A. Case Study*

The Ground Truth map used in the experiments, of the two algorithms aforesaid, is Acquired from AVIRIS sensor (AVIRIS92AV3C) [6], it contains 220 images. Each band has the size of 145x145 pixels, two thirds of this image are covered by agricultural land and one third by forests. Pixels are labeled as one of the 16 vegetation classes or unidentified. See figure 3.

Due to the availability of reference data, this hyperspectral image is an excellent source for the realization of experimental studies. 50% labeled pixels are randomly selected to be used in training, and the other 50% will be used for the classification testing [7]. The classifier used is SVM [8] [9] [10].

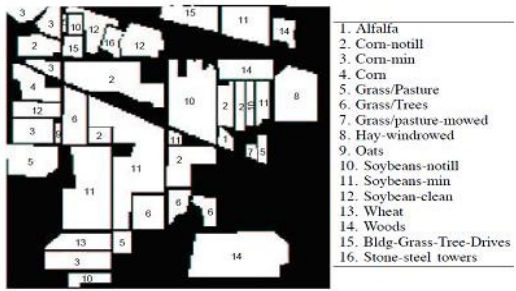

Fig. 3. The Ground Truth map of AVIRIS 92AV3C

*B. Results*

- The figure 4 illustrate the mutual information of the AVIRIS with the Ground Truth map, where we can see for example the noisy bands: 155 or 220. This explain the necessity to reduce the dimensionality of the HSI and to eliminate some bands.

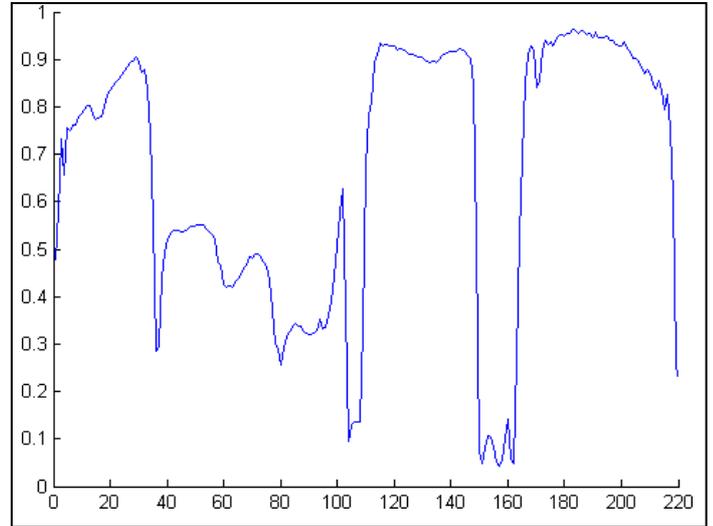

Fig. 4. Mutual information of AVIRIS with the Ground Truth map

- The plot of homogeneity feature of the 220 bands is shown in figure 5. By this statistic, we can also allocate the no informative bands affected by atmospheric effects for example, so we have to reduce dimensionality.

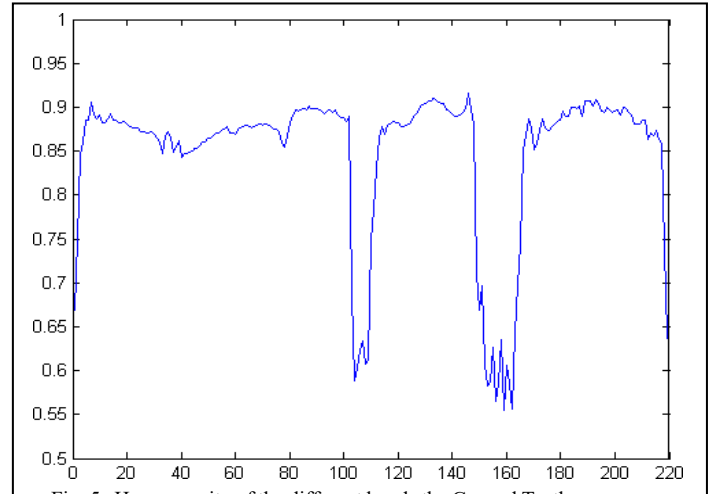

Fig. 5. Homogeneity of the different bands the Ground Truth map

- Now, we will apply the algorithms of process1 and process2, presented in the previous section, on the AVIRIS92AV3C to reduce the dimensionality and its classification.

The table I gives the classification results using the selection by mutual information ( process1 ) for different thresholds Th to control redundancy.

TABLE I. RESULTS OF ALGORITHM 1: REDUNDANCY CONTROL FOR DIFFERENT VALUES OF THRESHOLD (TH)

| | | The accuracy (%) of classification for numerous thresholds | | | | | |
|---|---|---|---|---|---|---|---|
| | | *-0,0200* | *-0,0100* | *-0,0050* | *-0,0040* | *-0,0035* | *0,0000* |
| Number of retained Bands | 2 | 47,44 | 47,44 | 47,44 | 47,44 | 47,44 | 47,44 |
| | 3 | 47,87 | 47,87 | 47,87 | 47,87 | 47,87 | 48,92 |
| | 4 | 49,31 | 49,31 | 49,31 | 49,31 | 49,31 | |
| | 12 | 56,30 | 56,30 | 56,30 | 56,30 | 60,76 | |
| | 14 | 57,00 | 57,00 | 57,00 | 57,00 | 61,80 | |
| | 18 | 59,09 | 59,09 | 59,09 | 62,61 | 63,00 | |
| | 20 | 63,08 | 63,08 | 63,08 | 63,55 | | |
| | 25 | 66,12 | 64,89 | 64,89 | 65,38 | | |
| | 35 | 76,06 | 74,72 | 75,59 | | | |
| | 36 | 76,49 | 76,60 | 76,19 | | | |
| | 40 | 78,96 | 79,29 | | | | |
| | 45 | 80,85 | 81,01 | | | | |
| | 50 | 81,63 | 81,12 | | | | |
| | 53 | 82,27 | 86,03 | | | | |
| | 60 | 82,74 | 85,08 | | | | |
| | 70 | 86,95 | | | | | |
| | 75 | 86,81 | | | | | |
| | 80 | 87,28 | | | | | |
| | 83 | 88,14 | | | | | |

The following table presents the different results obtained by using the algorithm 2, we can see the effectiveness selection of this algorithm, and the positive effect of the use of the extracted information "the homogeneity".

TABLE II. RESULTS OF ALGORITHM 2: REDUNDANCY CONTROL BASED ON M I AND HOMOGENEITY FOR DIFFERENT VALUES OF TH

| | | The accuracy (%) of classification for numerous thresholds | | | | | |
|---|---|---|---|---|---|---|---|
| | | *-0,0200* | *-0,0100* | *-0,0050* | *-0,0040* | *-0,0035* | *0,0000* |
| Number of retained Bands | 2 | 50,94 | 50,94 | 50,94 | 50,94 | 51,88 | 51,88 |
| | 3 | 55,85 | 55,85 | 55,85 | 55,85 | 52,56 | 52,56 |
| | 4 | 56,63 | 56,93 | 56,93 | 56,93 | 53,28 | |
| | 12 | 63,90 | 65,11 | 65,11 | 65,11 | 61,56 | |
| | 14 | 66,45 | 66,53 | 66,04 | 63,24 | 61,97 | |
| | 18 | 68,19 | 68,69 | 68,81 | 68,21 | 61,97 | |
| | 20 | 69,49 | 69,45 | 69,08 | 68,07 | | |
| | 25 | 70,80 | 73,43 | 69,84 | 68,91 | | |
| | 35 | 77,05 | 75,78 | 69,84 | | | |
| | 36 | 77,69 | 76,17 | 69,84 | | | |
| | 40 | 78,04 | 76,66 | | | | |
| | 45 | 81,01 | 76,66 | | | | |
| | 50 | 80,75 | 76,66 | | | | |
| | 53 | 80,81 | 76,66 | | | | |
| | 60 | 81,71 | 76,66 | | | | |
| | 70 | 84,96 | | | | | |
| | 75 | 83,42 | | | | | |
| | 80 | 83,42 | | | | | |
| | 83 | 83,42 | | | | | |

## C. Analysis and discussion

According to the tables, we can see that:

To increase the rate of classification, we allowed some redundancy by using negatives thresholds.

For high thresholds values, in the range of (-0.0035 to 0), few bands are selected because there is no redundancy, for example, with Th=0, just two bands are retained and the second method prevails because the accuracy is better than using the first one.

If we allowed some redundancy, for medium thresholds (-0.01 to -0.004), the classification rate increases. This is a very interesting region where we note a good behavior of the two methods specially the second, by using the homogeneity feature of bands, for less than 35 bands.

Now if we permit more redundancy, it's the case of Th= -0.02, we obtain the same accuracy with more retained bands, and we can't have interesting results for the reason that the redundancy becomes useless as it appears in table2 for more than 70 bands.

Finally, these simulations shows that both of the features selection methods, described in the previous section, give a good results but each one has its particularity. For example if the number of retained bands is less than 36, the combination of spectral and spatial features " algorithm2" prevails.

## VI. CONCLUSION

In this paper, we presented the problem of inefficiency and the redundancy of attributes in Hyperspectral images and the necessity to reduce their dimensionality by saving their propriety regarding to the multispectral images. For this, reduction dimensionality methods has been illustrated into two categories: by selection or feature extraction. In the first step we reproduced an algorithm based on feature selection byusing mutual information to select bands able to classify the pixels of the Grouth truth, then we proposed a second algorithm that integrate the homogeneity feature with the mutual information. The use of homgeneity allows to improve results for some values of thresholds. These proposed processes are a Filter strategy and were applied by using Hyperspectral dataset AVIRIS 92AV3C.

The selection was be effectively done to reduce dimentionality and classify the HIS, we can say that the the useful redundancy was conserved by using several thresholds.

The simplicity of theses algorithms allows them to be used for fast applications with medium performances and to be investigated and improved.